\documentclass[10pt,twocolumn,letterpaper]{article}

\usepackage{iccv}
\usepackage{times}
\usepackage{epsfig}
\usepackage{graphicx}
\usepackage{amsmath}
\usepackage{amssymb}
\usepackage{booktabs}
\usepackage{multirow}
\usepackage{multirow}
\usepackage{amsfonts}
\usepackage{graphics}
\usepackage{algorithm}
\usepackage{algorithmic}
\usepackage{xcolor}

\newcommand{\jose}[1]{\textcolor{black}{#1}}
\newcommand{\myparagraph}[1]{\vspace{0.0em}\noindent\textit{#1}}

\usepackage[pagebackref=true,breaklinks=true,letterpaper=true,colorlinks,bookmarks=false]{hyperref}

\iccvfinalcopy 

\def\iccvPaperID{10774} 

\ificcvfinal\fi

\begin{document}

\title{An Unpaired Shape Transforming Method\\ for Image Translation and Cross-Domain Retrieval}

\author{Kaili Wang*$\dagger$ \\
\and
Liqian Ma* \\
\and
Jose Oramas$\dagger$\\
\and
Luc Van Gool*\\

\and
Tinne Tuytelaars* \\

\and 
* KU Leuven, ESAT-PSI ~~~~~~$\dagger$ University of Antwerp, imec-IDLab \\
}

\maketitle
\ificcvfinal\thispagestyle{empty}\fi

\begin{abstract}
We address the problem of unpaired geometric image-to-image translation. Rather than transferring the style of an image as a whole, our goal is to translate the geometry of an object as depicted in different domains while preserving its appearance characteristics.
Our model is trained in an unpaired fashion, i.e. without the need of paired images during training. 
It performs all steps of the shape transfer within a single model and without additional post-processing stages.
Extensive experiments on the VITON, CMU-Multi-PIE and our own 
FashionStyle datasets show the effectiveness of the method. In addition, we show that despite their low-dimensionality, the features learned by our model are useful to the item retrieval task.

\textit{*This manuscript is a pre-print currently under review at the Elsevier Journal Computer Vision and Image Understanding.}

\end{abstract}

\section{Introduction}
\label{intro}

\footnote{Kaili Wang and Liqian Ma contribute equally.}
Image-to-image translation (I2I) refers to the process of generating a novel image, which is similar to the original input image yet different in some aspects. Typically, the input and output images belong to different {\em domains}, with images in the same domain sharing a common characteristic, e.g. going from photographs to paintings~(\cite{neuralStyleTransfer}), from greyscale to color images~(\cite{ColorGAN}), or from virtual (synthetic) to real images~(\cite{T2Net}). Apart from direct applications~(\cite{SRGAN}), I2I has proven valuable
as a tool for data augmentation~(\cite{GAN4augmentation}) or to learn a representation for cross-domain image retrieval~(\cite{sketchBasedRetrieval}).

Traditionally, each image domain is characterized by a different appearance or {\em style}, and I2I is therefore sometimes referred to as {\em style transfer}~(\cite{neuralStyleTransfer}).
While the translation process may drastically change the appearance or style of the input image, 
the image semantics are to be preserved, i.e. both input and output should represent the same objects and scene. Moreover, in most works, also the image geometry, i.e. the shape of the objects and the global image composition, is preserved. We refer to this as the image {\em content}. 

Most methods for I2I build on top of Generative Adversarial Networks (GANs)~(\cite{GAN,DCGAN,LSGAN,WGAN}) and are data-driven. They learn a translation model from example images of the two domains. While most methods require paired examples~(\cite{pix2pix,YooPixelLevelTransfer16,Zhao_2018_CVPR}), some recent methods 
do not~(\cite{DCGAN,cycleGAN,anonymous2019unsupervised}). 
To constrain the complexity of the problem,
the training data is often restricted to a specific setting, e.g. close-ups of faces~(\cite{huang2017beyond,Zhao_2018_CVPR}), people~(\cite{ma2017pose,ma2017disentangled}), traffic scenes (\cite{trafficScenesGAN}), etc. 

\begin{figure}
\centering
\includegraphics[width=0.5\textwidth]{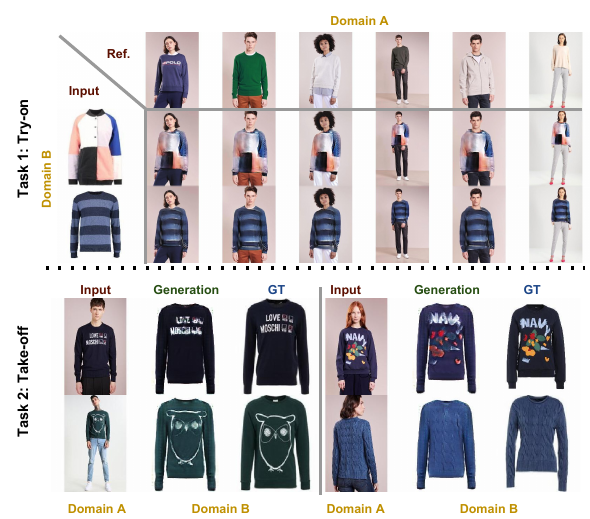}.
\caption{
Translating a clothing item from a "catalog" image domain to a domain of individuals wearing the indicated item (try-on task, top), and vice versa (take-off task, bottom). Notice how for both tasks the appearance details of the clothing items are preserved while their shape is effectively translated. 
}
\label{fig:TeaserImg}
\end{figure}

\begin{figure*}
\centering
\includegraphics[width=0.92\textwidth]{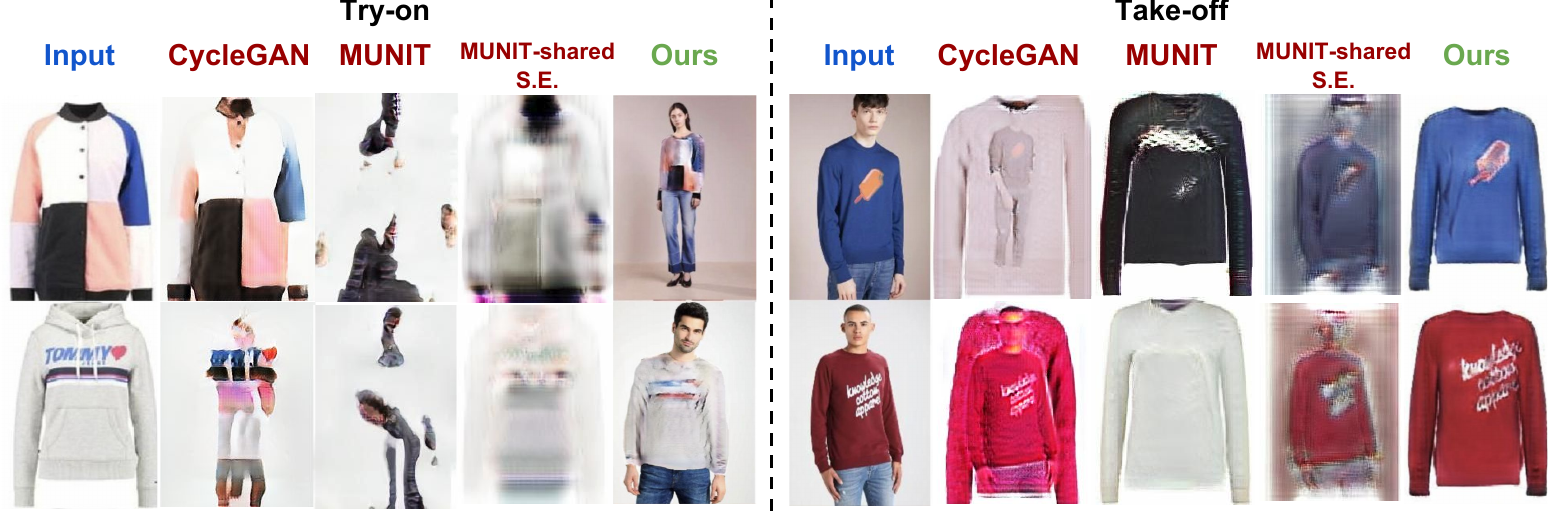}
\caption{Comparisons with CycleGAN (\cite{cycleGAN}) and MUNIT (\cite{MUNIT}) for try-on (left) and take-off (right) on FashionStyle dataset.
 }

\label{fig:baselineComparsion_intro}
\end{figure*}

In contrast to the traditional setting (\cite{ma2017disentangled,YooPixelLevelTransfer16,Zhao_2018_CVPR}), we focus on the challenge where input and output do {\em not} belong to  domains that share the same geometrical information. 
Instead, we work with one object-centric domain with standard shape and one that is more contextualized with 
large shape variation (using a reference image to provide the right context).
For instance, we go from a single piece of clothing to a person wearing that same item; or from a frontal face crop to a wider shot with arbitrary viewpoint of that same person (see Fig.~\ref{fig:TeaserImg} \& \ref{fig:faceQuality}). 
%
This setting is significantly more challenging, as the image geometry changes. At the same time, the image semantics (e.g. the clothing pattern or face identity) should be preserved. Analogous to the term style transfer, we refer to this as {\em shape transfer}.
While a couple of recent works have looked into this setting (\cite{ma2017disentangled,YooPixelLevelTransfer16,Zhao_2018_CVPR}), to the best of our knowledge we are the first to propose a solution that does {\em not} require paired data, across different domains, for model training. 
This is important, as collecting paired data is cumbersome or even impossible. 
Either way, it limits the amount of data that can be used for training, while access to large amounts of data is crucial for the quality of the results.  
Methods working with unpaired training data have been proposed for style transfer~(\cite{MUNIT,cycleGAN}), relying on low-level local transformations. However, these are not suited for the more challenging shape transfer setting, as clearly illustrated in Fig.~\ref{fig:baselineComparsion_intro}.

Translating shapes in a unpaired way is an unsolved task that is of interest for several reasons. 
First, it can be considered an alternative formulation of the novel-view synthesis problem, in the 2D image space, using only a single image as input. Second, shape translation can recover missing/occluded characteristics of an object instance which can help other tasks, such as recognition or tracking.

%
%
\jose{Beyond providing a wider comparison w.r.t. existing work, finer level of detail in the presentation, extended experiments and deeper discussions,
here we extend our recent work (\cite{wangICIP20}) along four directions. First, we conduct an ablation study highlighting the importance of the different components of the proposed method (Sec.~\ref{sec:translationAblationStudy}). Second, we perform an additional ablation study bridging the performance of our method w.r.t. similar methods from the literature, namely MUNIT~(\cite{MUNIT}) (Sec.~\ref{sec:ComponentsAblationStudy}). Third, we assess the capability of the proposed method at translating shapes across images of faces (Sec.~\ref{sec:exp_face}). Finally, we conduct a deep evaluation to assess the performance of the representations learned by the proposed method for the task of content-based image retrieval (Sec.~\ref{sec:exp_cloth_retrieval}).}

%
%
The main contributions of this paper are four-fold:
i) We analyze the task of 
\textit{\textbf{unpaired shape translation}}. 
To the best of our knowledge, we are the first doing this from an unpaired perspective.
%
ii) We propose an unpaired shape transforming (UST) method, which does not need any paired data or refinement post-processing. In one stream, an object with standard shape is transformed to a contextualized domain with arbitrary shape, and vice versa in the other stream.
iii) We achieve a one-to-many mapping by utilizing context and structure information guidance.
iv) 
We show the potential of the features learned by our model on the cross-domain item retrieval task.

This paper is organized as follows:
Sec.~\ref{sec:relatedWork} positions our work in the literature. In Sec.~\ref{sec:methodology} we present the details of the proposed method. This is followed by an extensive evaluation in Sec.~\ref{sec:experiment}. Finally, we draw conclusions in Sec.~\ref{sec:conclusion}.


\section{Related Work}
\label{sec:relatedWork}

\cite{pix2pix} first formulate the image-to-image translation problem with a conditional GAN model which learns a mapping from the source image distribution to the output image distribution using a U-Net neural network in an adversarial way. \cite{cycleGAN} propose  cycle-consistency to solve the I2I problem with unpaired data. This enables a lot of applications since it is usually expensive or even impossible to collect paired data for many tasks.~\cite{UNIT} assume that there exists a shared latent space for the two related domains and propose a weights-sharing based framework to enforce this constraint. These methods learn a one-to-one mapping function, \ie the input image is mapped to a deterministic output image. 
\cite{MUNIT, AugmentedCycleGAN, ma2018exemplar, lee2018diverse} propose unpaired multimodal methods which either sample multiple styles from a Gaussian space or capture the styles from exemplar images to generate diverse outputs.


All the above methods focus on appearance transfer where the content depicted in the input and output images has an aligned geometric structure.  \cite{Zhao_2018_CVPR,huang2017beyond,ma2017pose,ma2017disentangled,UnseenPose,SwapNet} aim at the case when the geometry itself is to be transferred. 
However, these methods focus on within-domain tasks (\eg person-to-person and face-to-face), which depicts reduced variability when compared to its cross-domain counterpart (\eg person-to-clothing).
Focused on images of clothing items,~\cite{YooPixelLevelTransfer16} propose one of the first methods addressing cross-domain pixel-level translation. Their method semantically transfers a natural image depicting a person (source domain) to a clothing-item image corresponding to the clothing worn by that person on the upper body (target domain), and vice versa. Recently, \cite{han2018viton, wang2018toward} propose two-stage warping-based methods aimed at virtual try-on of clothing items. These methods focus on learning a thin-plate spline (TPS) operation to transfer the pixel information directly. 
They rely on paired data to learn to transfer the shape in a first stage and then refine it in a second stage. 
In contrast, we propose a more general method that utilizes the context and shape guidance to perform translation across different domains without any paired data.
%
In addition, 
different from previous works which divide the translation 
process into multiple stages, our method is able 
to handle the full appearance-preserving translation, in 
both directions, within a single model. 

Outside of the I2I literature, \cite{spatialTransformerNet} proposes a spatial transformer network (STN) which also aims at object-level transformations. Different from our method, which learns the plausible transformations from data and allows for user-suggested transformations through the use of "desired" target images, STNs start from a predefined set of possible transformations. 
%
In addition, STNs apply 
the same transformation to every pixel. Differently, our method implicitly allows deformable objects since different pixel-level transformations are possible as depicted in the training data. 
Finally, STNs makes no distinction between content and style information.
Along the same line, recently \cite{lin2018stgan} and \cite{leeContextAwareSynthesis} proposed methods focused on the image composition task. \cite{lin2018stgan} learns how to add object segments with the correct shape in the semantic space. This is more constrained than our instance-level transfer in the RGB space. Moreover, it requires expensive supervision in the form of pixel-wise labels and instance-level contours. \cite{leeContextAwareSynthesis} operates in the RGB space. However, it can only handle homography transformations (in rigid objects) related to changes in viewpoint and scale. This is un-applicable on articulated/deformable objects and unsuitable to handle self-occlusions.

\section{Methodology}
\label{sec:methodology}

\begin{figure*}
  \centering
  \includegraphics[width=1\linewidth]{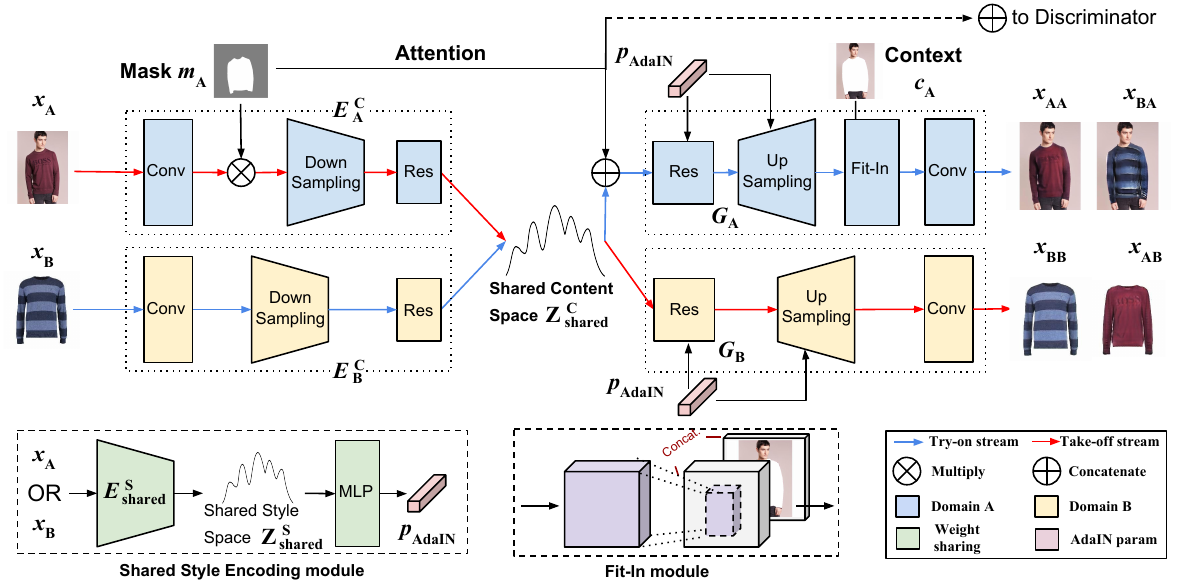}\\
  \caption{Proposed unpaired Shape Transforming (UST) method. 
  The try-on and take-off streams are trained jointly with shared style/content space constraints. 
To learn the one-to-many mapping in the try-on stream, the context information is utilized in the Fit-in module to constrain the output to be deterministic. Besides, an attention mechanism is applied to encourage the network to focus on the object. To learn the many-to-one mapping in the take-off stream, adversarial learning is adopted directly. 
  }
\label{fig:paper_framework}
\end{figure*}

In this section, we describe our model using the clothing try-on / take-off as an example. It should be noted though that our method can also be applied to other types of data, such as the face try-on / take-off illustrated in Sec.~\ref{sec:experiment}.
Our goal is to transfer the shape information while keeping the appearance information, all trained without access to paired data. For this, we propose the asymmetric two-stream model shown in Fig.~\ref{fig:paper_framework}. 
The asymmetry reflects the fact that one of the two domains (domain B) is object-focused (e.g. catalog images of clothing items) while the other one (domain A) shows the objects in context (e.g. pictures of clothed persons). 
In the one-to-many try-on stream (blue arrows), we transfer from the object-focused to the contextualized setting. This requires synthesizing a new image, where the shape of the object is first altered after which it is merged seamlessly with the provided context (in our setting, a segmented image of a person wearing a different piece of clothing). During this process, the color, texture and anything else specific to the object instance is to be preserved.  
In the many-to-one take-off stream (red arrows), our goal is to synthesize the clothing catalog image in a standard frontal view starting from a natural person image with varying pose. 

Here, we use $x_A$ and $x_B$ to refer to images from domain A and domain B
respectively. $x_{AB}$ refers to images transferred from domain A to domain B,
and vice versa for $x_{BA}$.

\subsection{Assumptions}
In previous works~(\cite{YooPixelLevelTransfer16,han2018viton}), the try-on and take-off tasks are solved in a supervised way, respectively. Here, we solve both tasks in one model using unpaired data based on 
shared-latent space and context-structure constraints.
%
%

\myparagraph{Shared-latent space constraint.}
Similar to~\cite{MUNIT,ma2018exemplar,lee2018diverse}, we decompose the latent space into a content space and a style space. Different from previous works, we have two assumptions: 1) content space constraint, \ie the content space {\em can be shared} by the two domains; 2) style space constraint, \ie  images from the two domains {\em do share} the same style space. 
%
We use $\text{Z}_\text{A}^\text{C}$ and $\text{Z}_\text{B}^\text{C}$ to denote the content space of domains A and B, respectively. We assume $\text{Z}_\text{A}^\text{C}$ and $\text{Z}_\text{B}^\text{C}$ are both embedded in a larger latent space $\text{Z}_\text{shared}^C$.
Symbols $\text{Z}_\text{A}^\text{S}$ and $\text{Z}_\text{B}^\text{S}$ denote the style space of domain A and B, respectively. Note that we assume $\text{Z}_\text{A}^\text{S}$ and $\text{Z}_\text{B}^S$ are the same space, which is a stronger constraint.

To achieve the content space constraint, we use two encoders $E_\text{A}^\text{C}$ and $E_\text{B}^\text{C}$ to encode images from domain A and B, respectively. Then, we use a latent content code reconstruction loss 
to enforce the latent content reconstruction, similar to \cite{MUNIT,lee2018diverse}.
To achieve the style space constraint, we utilize both the weight-sharing technique~(\cite{UNIT}) and the latent style code reconstruction loss (see Fig.~\ref{fig:paper_framework}). 

\myparagraph{Context constraint.}
Although the above shared-latent space constraints enable the unpaired I2I and work well for 
style 
transfer tasks~(\cite{UNIT,ma2018exemplar,lee2018diverse}), it is not enough for geometry transfer when the output is multi-modal (\ie multiple possible outputs). To address this,  \cite{YooPixelLevelTransfer16} proposed triplet adversarial learning with paired data. However, for the unpaired setting, the adversarial learning on its own is too weak (see Fig.~\ref{fig:baselineComparsion_intro}). Here, we propose to use context information guidance to constrain the output to be deterministic, \ie decompose the one-to-many mapping into one-to-one mappings. In particular, for the try-on stream, we propose a Fit-in module which combines the feature maps with the context information. 
As to the take-off stream, we assume the output is unimodal and directly use the adversarial learning to learn the deterministic many-to-one mapping.

\myparagraph{Attention.} 
Since domain A is contextualized, we need to constrain the network to focus on the object instead of the background. Therefore, we introduce the attention mechanism in both generator and discriminator, \ie concatenating the shape mask $m_A$ with the inputs before $G_A$ and the discriminator $D_A$ (see Fig.~\ref{fig:paper_framework}).

\subsection{Network architecture}
The model can be divided into several sub-networks.
For the content encoder $E_\text{A}^\text{C}$ and $E_\text{B}^\text{C}$, we use a convolution block and several down-sampling layers followed by several residual blocks.
The decoders $G_\text{A}$ and $G_\text{B}$ are symmetric with the encoding part except for the Fit-in module which is key to learn the one-to-many mapping and the shape mask attention that helps preserving the appearance. The Fit-in module is a simple convolution block which receives 
the generated feature map and the context information of the desired target shape. 
%
%
%
The shared style encoding module contains a style encoder $E_\text{shared}^\text{S}$ and a multilayer perceptron (MLP). It encodes the style information of both domains.


\myparagraph{Try-on stream}
The catalog image $x_\text{B}$ first passes through the domain B content encoder $E_\text{B}^\text{C}$ 
producing the content code $z_\text{B}^\text{C}$ in the shared content space $\text{Z}^\text{C}_{shared}$. 
In parallel, $x_\text{B}$ is also encoded into a style code $z_\text{B}^\text{S}$ in the shared style space $\text{Z}_\text{shared}^\text{S}$ by the shared style encoder $E_{shared}^S$. 
To combine the content and style information in the decoder, we use adaptive instance normalization (AdaIN, \cite{huang2017arbitrary}) layers for all residual and up-sampling blocks. 
%
The AdaIN parameters $p_\text{AdaIN}$ are dynamically computed by a multilayer perceptron from the style code $z_\text{B}^\text{S}$ to ensure the generated person image $x_\text{BA}$ has the same style as $x_\text{B}$.
\begin{align}
    \mathrm{AdaIN}(z,\gamma,\beta) = \gamma \frac{(z-\mu(z))}{\delta(z)} + \beta
    \label{eq:AdaIN},
\end{align}
where $z$ is the activation of the previous convolution layer. $\mu$ and $\delta$ are the mean and standard deviation computed per channel. Parameters $\gamma$ and $\beta$ are  the output of the MLP of the shared style encoding module. 


During decoding, the content code $z_\text{B}^\text{C}$ concatenated with the shape mask $m_\text{A}$ are fed to the decoder $G_\text{A}$. There the content and style are fused by AdaIN and then fed to the Fit-in module. We apply AdaIN in both the residual blocks and up-sampling layers. This helps stabilize and speed up the convergence during training, and also helps preserve appearance better. This is due to the fact that AdaIN can be treated as a skip-connection between the encoder and decoder to alleviate the exploding and
diminishing gradient problems.
The Fit-in module is designed to enforce the context information constraint. 
We first obtain the bounding box of the mask from the context image. Then, we resize and align the up-sampled feature maps to this bounding box. Finally, this ouput is concatenated with the context image.
The main goal of this design is to integrate the context information which helps the deterministic shape transform. The final try-on image $x_\text{BA}$ is generated after the last convolution block.

In addition, inspired by~\cite{pix2pix}, we introduce an attention mechanism to both generator and discriminator. We concatenate
the mask $m_\text{A}$ with the content code $z_\text{B}^\text{C}$ before the generator $G_A$ and concatenate the mask $m_\text{A}$ with the generated image $x_\text{BA}$ before the discriminator $D_A$, respectively.
This simple but effective attention operation encourages the network to focus on the generated clothing instead of the context part. 
This improves the results, especially when the objects to be translated have a highly variable scale/location within the images. 


\myparagraph{Take-off stream}
For the take-off stream, the clothed person image $x_\text{A}$ first passes through a convolution block and then gets multiplied with the clothing mask $m_\text{A}$ in order to exclude the background and skin information. Similar to the try-on stream, the masked feature maps are then encoded into a content code $z_{\text{A}}^\text{C}$ in the shared content space $\text{Z}_\text{shared}^\text{C}$. 

For the decoding part, the only difference with the try-on stream is that there is no "Fit-in" module or mask attention. The final take-off catalog image $x_\text{AB}$ is generated by decoding $z_{\text{A}}^\text{C}$,
through the decoder $G_\text{B}$ with AdaIN residual blocks, up-sampling blocks and convolution blocks. 


\subsection{Learning}
\label{sec:opt}
In this section, we only describe $\text{A} {\rightarrow} \text{B}$ translation for simplicity and clarity. 
The $\text{B} {\rightarrow} \text{A}$ is learned in a similar fashion.
We denote the content latent code as 
{$\text{z}_\text{A}^\text{C}{=}E_\text{A}^\text{C}(x_\text{A})$}, 
style latent code as 
{$\text{z}_\text{A}^\text{S}{=}E_\text{shared}^\text{S}(x_\text{A})$}, 
within domain reconstruction output as 
{$x_\text{AA} {=} G_\text{A}(\text{z}_\text{A}^\text{C},\text{z}_\text{A}^\text{S})$}, 
cross domain translation output as 
{$x_\text{AB} {=} G_\text{B}(\text{z}_\text{A}^\text{C},\text{z}_\text{A}^\text{S})$}. 
Our loss function contains terms for the bidirectional reconstruction loss, cycle-consistency loss and adversarial loss~\cite{MUNIT,lee2018diverse}. Besides, we also use a composed perceptual loss to preserve the appearance information across domains, and a symmetry loss capturing some extra domain knowledge~(\cite{huang2017beyond,Zhao_2018_CVPR}).

\myparagraph{Bidirectional reconstruction loss ($L^{x_A}_{LR}$, $L^{x_A}_{SR}$).}
This loss consists of the feature level latent reconstruction loss $\mathcal{L}_{\text{LR}}$ and the pixel level image self-reconstruction loss $\mathcal{L}_{\text{SR}}$. The former  contains both content and style code reconstructions. The whole bidirectional reconstruction loss encourages the network to learn encoder - decoder pairs that are inverses of one another and also stabilizes the training.
\begin{align}
    \scriptstyle
    \mathcal{L}_{\text{LR}}^{x_\text{A}} = &\mathbb{E}_{x_\text{AB},z_\text{A}^\text{C}}\big[\|E_\text{B}^\text{C}(x_\text{AB}) - \text{z}_\text{A}^\text{C}\|_1\big] \nonumber \\
    + &\mathbb{E}_{x_\text{AB},z_\text{A}^\text{S}}\big[\|E_\text{shared}^\text{S}(x_\text{AB}) - \text{z}_\text{A}^\text{S}\|_1\big] \\
    \scriptstyle
    \mathcal{L}_{\text{SR}}^{x_\text{A}} = &\mathbb{E}_{x_\text{A}}\big[\|x_\text{AA}-x_\text{A}\|_1\big],  
    \label{eq:IR_loss}
\end{align}

%
%
%
%
\myparagraph{Adversarial loss ($L^{x_A}_{GAN}$).}
To make the translated image look domain realistic, we use an adversarial loss to match the domain distribution. For the $\text{A} {\rightarrow} \text{B}$ translation, the domain B discriminator $D_B$ tries to distinguish the generated fake images with the real domain B images, while the generator $G_B$ will try to generate domain B realistic images.
\begin{align}
\scriptstyle
\mathcal{L}_{GAN}^{x_\text{A}} = {\mathbb{E}}_{x_\text{B}}\big[\log{D_\text{B}(x_\text{B})}\big] + {\mathbb{E}}_{x_\text{AB}}\big[\log{(1-D_\text{B}(x_\text{AB}))}\big]
    \label{eq:GAN_loss}
\end{align}
\myparagraph{Cycle-consistency loss ($L^{x_A}_{CC}$).}
To enable unpaired translation, the cycle-consistency loss~\cite{cycleGAN} is applied to stabilize the adversarial training.
\begin{align}
    \scriptstyle
    \mathcal{L}_{\text{CC}}^{x_\text{A}} = \mathbb{E}_{x_\text{AB},x_\text{A}}\big[\|G_\text{A}(E_\text{B}^\text{C}(x_\text{AB}),E_\text{shared}^\text{S}(x_\text{AB})) -x_\text{A}\|_1\big]
    \label{eq:CC_loss}
\end{align}
\myparagraph{Perceptual loss ($L^{x_A}_P$).}
To preserve the appearance information, we apply a composed perceptual loss.
%
%
\begin{align}
    \scriptstyle
    \mathcal{L}_{\text{P}}^{x_\text{A}} = &(\mathbb{E}_{x_\text{AA},x_\text{A}}\big[\|\Phi(x_\text{AA})-\Phi(x_\text{A})\|_2^2\big]) \nonumber \\
   + &(\mathbb{E}_{x_\text{AB},x_\text{B}}\big[\|\Phi(x_\text{AB})-\Phi(x_\text{A'})\|_2^2\big]) \nonumber \\
   + & \lambda~\mathbb{E}_{x_\text{AB},x_\text{B}}\big[\|Gram(x_\text{AB})-Gram(x_\text{A'})\|_1\big],
    \label{eq:perceptual_loss}
\end{align}

%
%
%
where $x_{A'}$ is the Region of Interest (RoI) of $x_A$. For clothing items, it is
the segmented clothing region. For the face experiments, it is the facial region (without context information).
$\Phi$ is a network trained on external data, whose representation can capture image similarity.
Similar to~\cite{gatys2016image} and \cite{johnson2016perceptual}, we use the first convolution layer of all five blocks 
in VGG16~{\cite{VGG}} to extract the feature maps to calculate the Gram matrix which contains non-localized style information.
$\lambda$ is the corresponding loss weight.

\myparagraph{Symmetry loss ($L^{x_A}_{Sym}$).} To utilize the inherent prior knowledge of clothing and human faces, we apply a symmetry loss~(\cite{huang2017beyond,Zhao_2018_CVPR}) to the take-off stream.
\begin{align}\label{eq:Sym_loss}
    \scriptstyle
    \mathcal{L}_{\text{Sym}}^{x_\text{A}}=
    \mathbb{E}_{x_\text{AB}}\big[
    \frac{1}{W/2\times H}\!\sum_{w=1}^{W/2} \sum_{h=1}^{H}
    \|x_\text{AB}^{w,h}-x_\text{AB}^{W-(w-1),h}
    \|_1\big],
\end{align}
where $H$ and $W$ denote the height and width of the image, $(w,h)$ 
are the coordinates of each pixel, and $x_\text{AB}^{w,h}$ refers 
to a pixel in the transferred image $x_\text{AB}$.

\myparagraph{Total loss.}
Our model, including encoders, decoders and discriminators, is optimized jointly. The full objective is as follows,
\begin{align}
\scriptstyle
&\min\limits_{\substack{E_\text{A}^\text{C},E_\text{B}^\text{C},E_\text{shared}^\text{S},\\G_\text{A},G_\text{B}}}\max\limits_{D_\text{A},D_\text{B}}
    \mathcal{L}(E_\text{A}^\text{C},E_\text{B}^\text{C},E_\text{shared}^\text{S},G_\text{A},G_\text{B},D_\text{A},D_\text{B}) \nonumber \\
    &= \mathcal{L}_{\text{GAN}}^{x_\text{A}}+\mathcal{L}_{\text{GAN}}^{x_\text{B}}
    +\lambda_\text{CC}(\mathcal{L}_{\text{CC}}^{x_\text{A}}+\mathcal{L}_{\text{CC}}^{x_\text{B}})
    +\lambda_\text{SR}(\mathcal{L}_{\text{SR}}^{x_\text{A}}+\mathcal{L}_{\text{SR}}^{x_\text{B}})  \nonumber \\
    &+\lambda_\text{LR}(\mathcal{L}_{\text{LR}}^{x_\text{A}}+\mathcal{L}_{\text{LR}}^{x_\text{B}})
    +\lambda_\text{P}(\mathcal{L}_{\text{P}}^{x_\text{A}}+\mathcal{L}_{\text{P}}^{x_\text{B}}) +\lambda_\text{Sym}\mathcal{L}_{\text{Sym}}^{x_\text{A}}.
    \label{eq:total_loss}
\end{align}
where $\lambda_\text{CC}$, $\lambda_\text{SR}$, $\lambda_\text{LR}$, $\lambda_\text{P}$ and $\lambda_\text{Sym}$ are loss weights for different loss terms.

\section{Evaluation}
\label{sec:experiment}

We evaluate our method on both clothing try-on / take-off and face 
try-on / take-off tasks. We perform an ablation study on our own 
FashionStyle dataset. 
Then, the full model results on VITON and MultiPIE datasets are reported.
Finally, we assess the potential of the learned 
style/appearance representation for clothing item 
retrieval across domains.

\myparagraph{Datasets}
We use three datasets: VITON (\cite{han2018viton}), Fashion-Style and CMU MultiPIE~(\cite{MultiPIE}).
VITON and FashionStyle are fashion related datasets, see Figs.~\ref{fig:TeaserImg}, \ref{fig:FashionStyle_ABL}, \ref{fig:VITONQuanlitative} for some example images.
VITON has around 16,000 images for each domain.
However, we find that there are plenty of image duplicates with different
file names. After cleaning the dataset, there are 7,240 images in each domain left.
The FashionStyle dataset, provided by an industrial partner, has 5,230 training images and 1,320 testing images of clothed people (domain A), and 2,837 training images and 434 testing images of the clothing catalog items (domain B). For domain A, FashionStyle has multiple views of the same person wearing the same clothing item. %
We present results on this dataset for one category, namely pullover/sweater. 
CMU MultiPIE is a face dataset under pose, illumination and expression changes. Here we focus on a subset of images
with neutral illumination and expression, 
and divide the subset in two domains: 7,254 profile images (domain A) and 920 frontal views (domain B). 


\myparagraph{Metrics} We use paired images from different domains depicting the same clothing item to quantitatively evaluate the performance our method. 
For the case of the try-on task we measure the similarity 
between the ROI of original image (from domain A) and the ROI of a generated version
(where its corresponding clothing item has been translated to fit in a 
masked out version of the image). Thus, we call it \textit{Try-on ROI}. 
To create this masked image we first 
run a clothing-item segmentation algorithm~(\cite{liang2018look}) that we 
use to remove the clothing-item originally worn by the person.
For the case of the take-off task, given an image from domain A, 
we measure the similarity of its corresponding clothing item (from domain B) 
with the generated item.
On both cases similarity between images is computed using the SSIM (\cite{SSIM}) and LPIPS~(\cite{zhang2018unreasonable}) metrics. 
We report the mean similarity across the whole testing set.


For the retrieval task performance is reported in terms of Recall rate given that in our dataset every query image has only one corresponding image in the database.


\myparagraph{Implementation details}
The perceptual feature extractors $\Phi$ in Eq.~\ref{eq:perceptual_loss} are LPIPS~(\cite{zhang2018unreasonable}) and Light-CNN~(\cite{wu2018light}) networks for clothing translation and face translation, respectively. 
In all our experiments, we use the Adam~\cite{Adam} optimizer with $\beta_1{=}0.5$ and $\beta_2{=}0.999$. The initial learning rate is set to $2 {\times} 10^{-6}$. 
Models are trained with a minibatch of size $1$ for FashionStyle and VITON, and $2$ for the face experiment. We use the segmentation method~\cite{liang2018look} 
to get the clothing mask and its bounding box. For faces, we detect the face landmarks using the detector proposed by \cite{cao2017realtime} and then connect each point to get the face mask.
The shared content code is a tensor whose dimension is determined by the data.
The shared style code is a vector, we use $8/32/128$ dimensions in our experiments.

Table \ref{tab:param} summarizes the details of network architecture presented in Fig. 3 of the submitted manuscript. Furthermore, we specify the training parameters used for the tasks analyzed in our experiments. The number of input/output convolution blocks is set to $n_{1} = 1$. The number of down-sampling and up-sampling convolution blocks is set to $n_{2} = 3$ and $n_{2} = 2$ for clothing and face translation, respectively. We need a different value here, since the images from the two datasets have different resolutions. The number of residual blocks is set to $n_{3} = 4$ for both clothing and face translation experiments. As for our Fit-in module, it consists of one residual block to merge the features with the context information.

\begin{table}[h]
\centering

\caption{Network architecture and training parameters details.}
\resizebox{1\columnwidth}{!}{
\begin{tabular*}{12cm}
{@{\extracolsep{\fill}} l | c c | c c}
\toprule 
\multirow{2}{*}{Parameter} & Clothing & Clothing & Face & Face  \\ 
                        & try-on & take-off & try-on & take-off\\

\midrule[0.6pt]	
	$n_1$  &1 &1 &1 &1\\
	$n_2$  &3 &3 &2&2\\
	$n_3$ &4 &4 &4&4 \\
	Minibatch & 1 & 1&2&2 \\
	Learning rate & 4e-5 & 4e-5 &4e-5&4e-5\\ 
	$\lambda_{CC}$ & 5 & 5 &5&5\\ 
	$\lambda_{SR}$ & 10 & 10 &10&10\\ 
	$\lambda_{LR}$ & 10 & 10 &10&10\\
	$\lambda_{Sym}$ & 10 & 10 &10&10\\
	$\lambda_P$ & 5 & 2.5 &0.075&0.025\\
	$\lambda$ & 2e4 & 2e4 &6e5&2e6\\
	Iteration &$\sim$60k &$\sim$60k &$\sim$60k&$\sim$60k\\

\bottomrule[1pt]

\end{tabular*}
}
\label{tab:param}
\end{table}



\begin{figure*}[thb]
\centering
\includegraphics[width=1\textwidth]{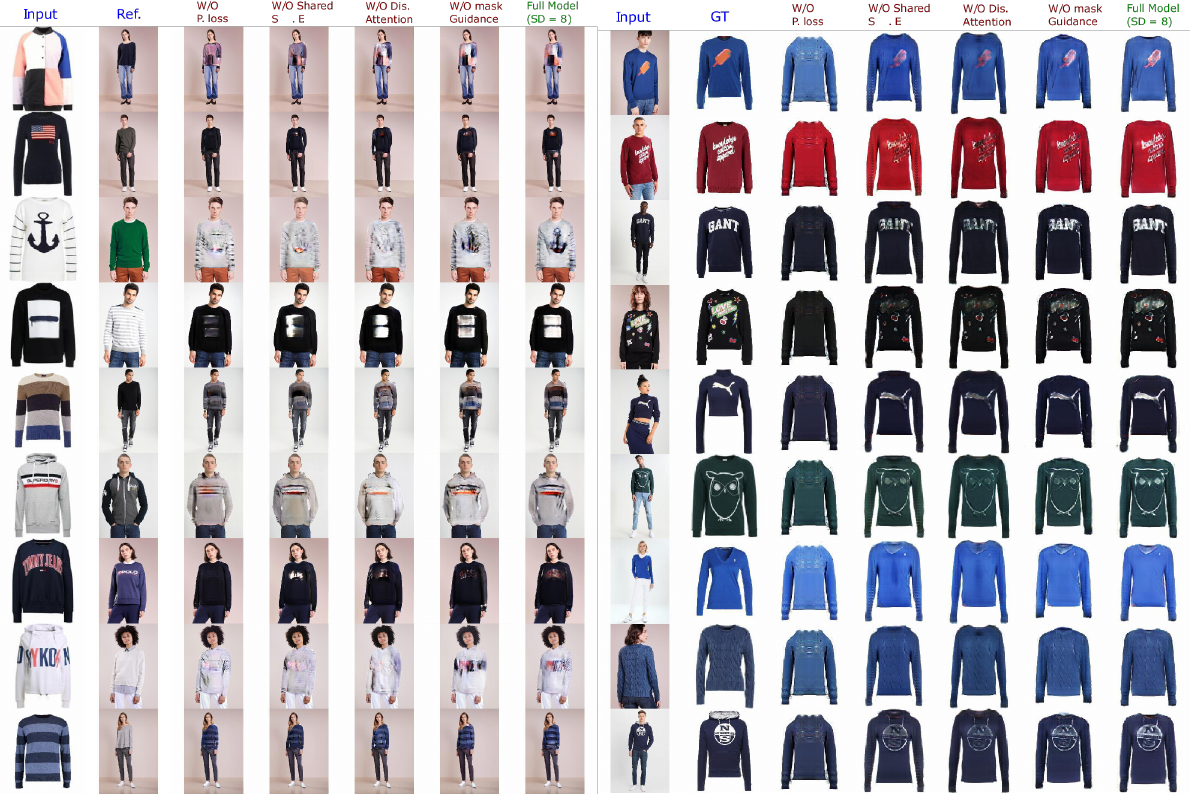}
\caption{
Ablation study on the FashionStyle dataset. 
The top part are try-on results and the bottom part are take-off results. The first two columns show the input image and the reference ground truth image. The other columns show the generated results of different model settings. 
Please zoom in for more details. 
}
\label{fig:FashionStyle_ABL}
\end{figure*}


\begin{figure*}[thb]
\centering
\includegraphics[width=1\textwidth]{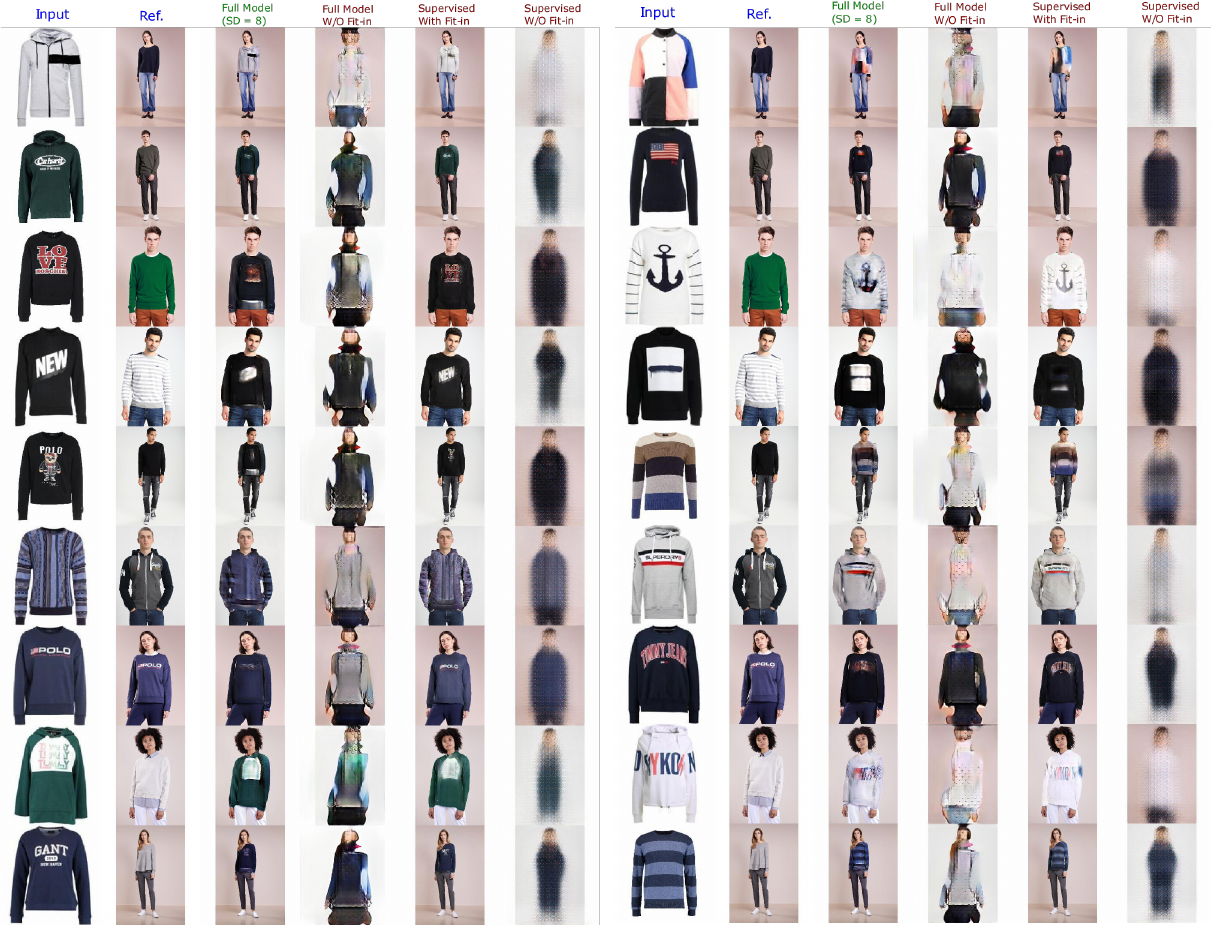}
\caption{Ablation study on FashionStyle Dataset w.r.t. the Fit-in Module. It clearly shows without the Fit-in module, even the supervised model can not generate the sharp, clean images.  
}
\label{fig:FashionStyle_ABL}
\end{figure*}


\begin{figure*}[thb]
\centering
\includegraphics[width=1\textwidth]{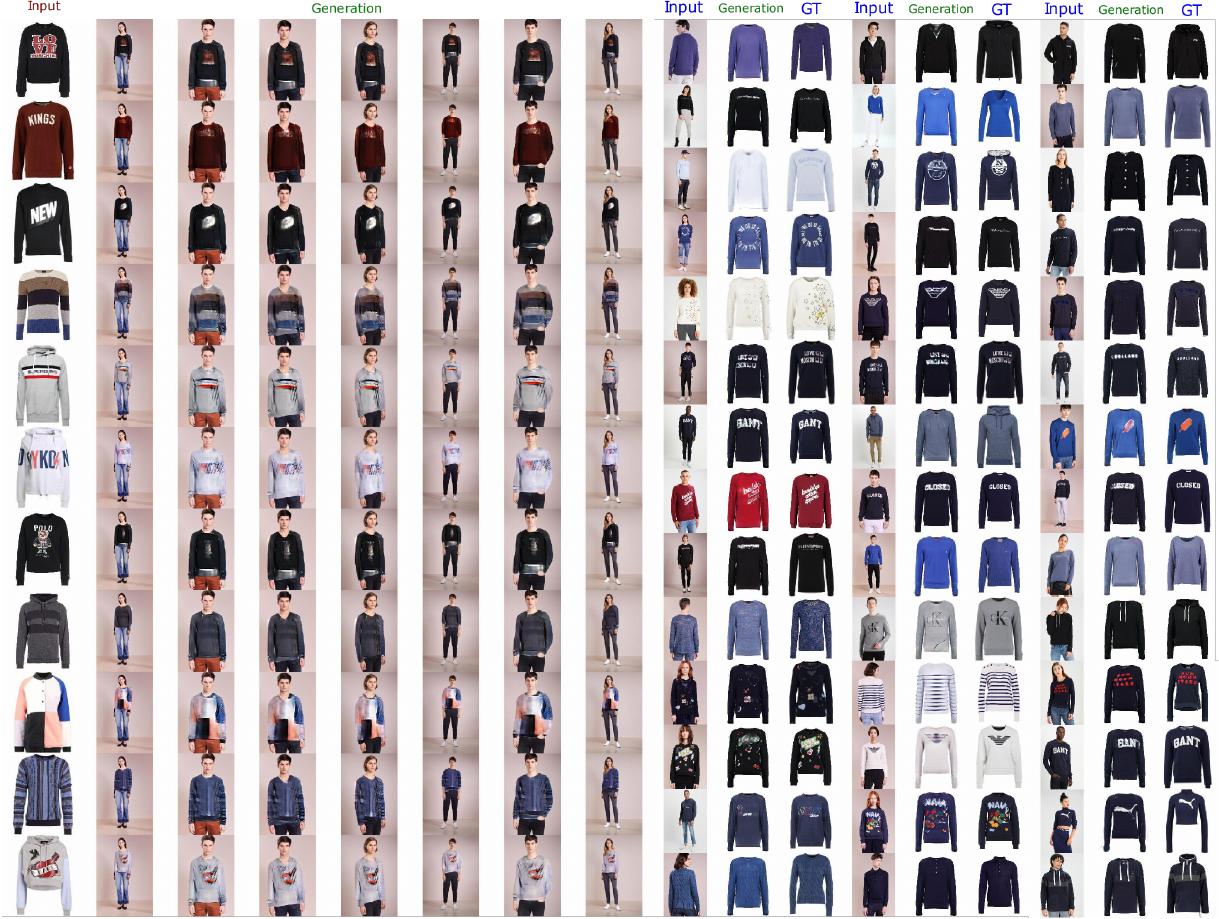}
\caption{The quality results of try-on (left) and take-off (right) tasks. Please note the "GT" in the take-off is just a reference.
The whole model is trained by using unpaired data.
}
\label{fig:FashionStyle_ABL}
\end{figure*}



\begin{figure}[htb]
\centering
\includegraphics[width=0.5\textwidth]{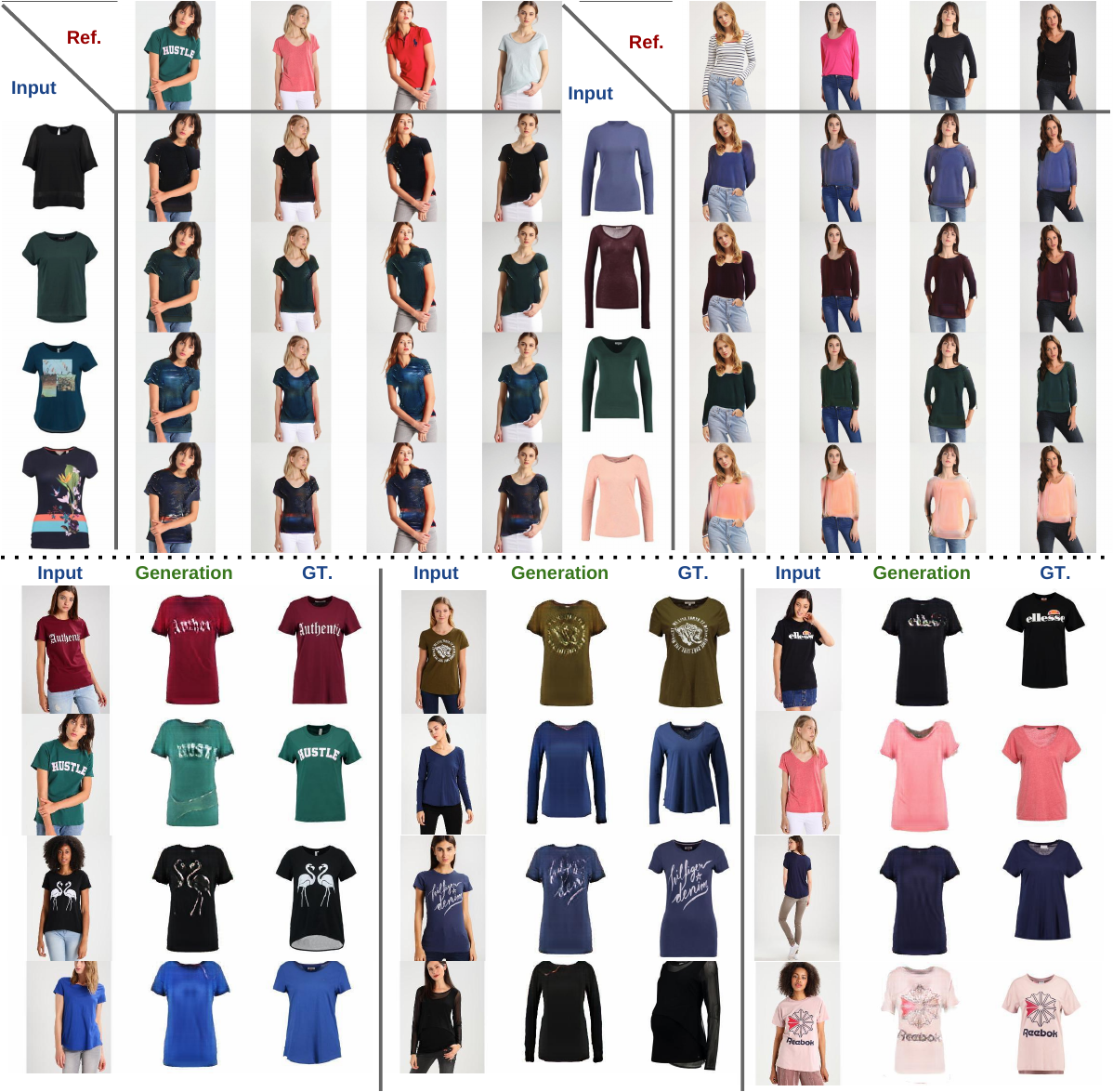}
\caption{
     Try-on and take-off results on the VITON dataset. For try-on (top) each column shows a person (from the top row) virtually trying on different clothing items. For take-off (bottom) each example consists of three images: input image, generated take-off image and
the ground-truth (GT) image.  Zoom in for more details. 
}
\label{fig:VITONQuanlitative}
\end{figure}


\subsection{Ablation Study: Clothing try-on / take-off}
\label{sec:translationAblationStudy}
%

We conduct a study in order to analyze the importance of four main 
components of our model. More precisely, the \textit{perceptual loss}, shared style encoder (\textit{Shared S. E.}), 
\textit{mask attention}
and \textit{Fit-in module}, on the FashionStyle dataset. 
Note that \textit{mask attention} is applied to generator $G_A$ and discriminator $D_A$.
%
Towards this goal we test different variants of our architecture 
(Sec.~\ref{sec:methodology}) where one of these four components has been 
removed.
In addition, we run an experiment using a \textit{supervised model} (paired data).
The model architecture 
is a residual block based on U-net similar to PG$^2$~(\cite{ma2017pose}), but 
extended to get closer to our model.
It is extended by applying our mask multiplication operation after the first convolution 
block for the supervised take-off experiment. Likewise, we add our Fit-in module 
for the supervised try-on experiment. 
%
We present quantitative results on the translation performance of the try-on / take-off
tasks in Table~\ref{tab:FashionStyle_similarity} for the FashionStyle dataset
with related qualitative results presented in Fig.~\ref{fig:FashionStyle_ABL}.


\begin{table}
\setlength{\tabcolsep}{4.7pt} 
\centering
\caption{Mean SSIM and LPIPS-VGG similarity of each setting from our ablation study. 
Higher SSIM values and lower LPIPS indicate higher similarity. 
Both metrics are in the range $[0,100]$.
}
\resizebox{1\columnwidth}{!}{
\begin{tabular*}{12cm}
{@{\extracolsep{\fill}} l c c }
\toprule 
\multirow{2}{*}{Method} & Try-on ROI  & Take off \\ 
                        & (SSIM/LPIPS-VGG) & (SSIM/LPIPS-VGG)\\

\midrule[0.6pt]	
W/O P. Loss & \textbf{66.78} / 27.37  & 58.96 / 36.62   \\
W/O shared S.E. & 66.72 / 27.38 & 60.33 / 34.94    \\
W/O mask attention & 64.63 / 28.13 & 60.94 / 34.49 \\
W/O Fit-in module & N/A / N/A & 60.11 / 37.64 \\
Full model & 66.42 / \textbf{27.02}& \textbf{61.19} / \textbf{34.37} \\
\midrule[0.3pt]	
Supervised model & \textit{\textbf{69.51}} / \textit{\textbf{24.14}} &\textit{\textbf{61.54}} / \textit{\textbf{32.56}} \\

\bottomrule[1pt]
\end{tabular*}
}

\label{tab:FashionStyle_similarity}
\end{table}


\myparagraph{Discussion.} 
A quick inspection of Table~\ref{tab:FashionStyle_similarity} reveals 
that, based on the LPIPS metric, the full model generates images with the highest similarity to the ground-truth on the try-on task among the 
unpaired variants. 
Our full model generates sharper and more consistent results than other models, but does  not obtain the highest SSIM. This is also observed in person generation and super-resolution papers~\cite{ma2017pose,johnson2016perceptual}.
The try-on ROI scores 
of W/O Fit-in module is not applicable since without the context information, the network cannot determine the target generated shape, i.e. ROI cannot be determined. 
This task seems to be affected most when the \textit{mask attention} is dropped.
This confirms the relevance of this feature when translating shape from images in this direction (try-on).

For the case of the take-off task, results are completely
dominated by the full model among the unpaired variants. However, different from the try-on task, the take-off task 
is mostly affected by the removal of the perceptual loss(i.e. LPIPS) and
Fit-in modules. The Fit-in module is set in the try-on stream, but since the two streams are trained jointly, the take-off stream is indirectly affected by the performance of the try-on stream. Therefore, the take-off result of W/O Fit-in module is the worst.
Although this trend is different from the try-on task, it is not 
surprising given that for the take-off task, the expected shape of the 
translated image is more constant when compared with that of the 
try-on task which is directly affected by the person's pose.
Moreover, the output of the take-off task is mostly dominated 
by uniformly-coloured regions, which is a setting in which 
perceptual similarity metrics, such as LPIPS, excel.

A close inspection of Fig.~\ref{fig:FashionStyle_ABL} confirms the trends previously observed. Note how the full model produces the most visually-pleasing result; striking a good balance between shape and level of details 
on the translated items.
The other unpaired variants
(except \textit{W/O Fit-in module})
tend to generate blurry results and lose details, e.g. patterns and logos, while maintaining the basic shape and context well.  
More critical, \textit{W/O Fit-in module} no longer preserves both shape and loses details.
Especially, without context information guidance, it is difficult for the model to learn the one-to-many mapping resulting in inconsistent outputs.
We have noted that failures are mostly caused by incorrectly estimated masks and heavy occlusion.

It is remarkable that quantitatively speaking 
(Table~\ref{tab:FashionStyle_similarity}), the performance 
of our method is comparable to that of the \textit{supervised model}.
Moreover, while the \textit{supervised model} is very good at translating 
logos, our method  still has an edge when translating patterns 
(e.g. squares from the $1^{st}$ row and stripes from the $3^{rd}$ row of Fig.~\ref{fig:FashionStyle_ABL}), without requiring paired data.

\subsection{Clothing try-on / take-off on VITON}

We complement the results presented previously  
with a qualitative experiment (see Fig.~\ref{fig:VITONQuanlitative})
on the VITON dataset using the full model.
We see that our method is able to effectively translate 
the shape of the clothing items across the domains.
It is notable that on the try-on task, it is able to preserve the texture information of the items even in the presence 
of occlusions caused by arms. This is handled by the proposed Fit-in module (Sec.~\ref{sec:methodology}) which learns how 
to combine foreground and contextual information.


\begin{figure}
\centering
\includegraphics[width=0.5\textwidth]{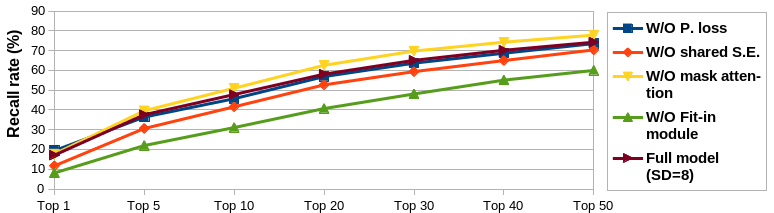}
\caption{Clothing retrieval ablation study. Note that relevant
factors for the retrieval are somewhat the opposite of those from
the translation task.
}

\label{fig:FashionStyle_Retri_ABL}
\end{figure}



\subsection{Comparisons with existing methods}
\label{sec:baselineComparsion}
\begin{figure}
\centering
\includegraphics[width=0.5\textwidth]{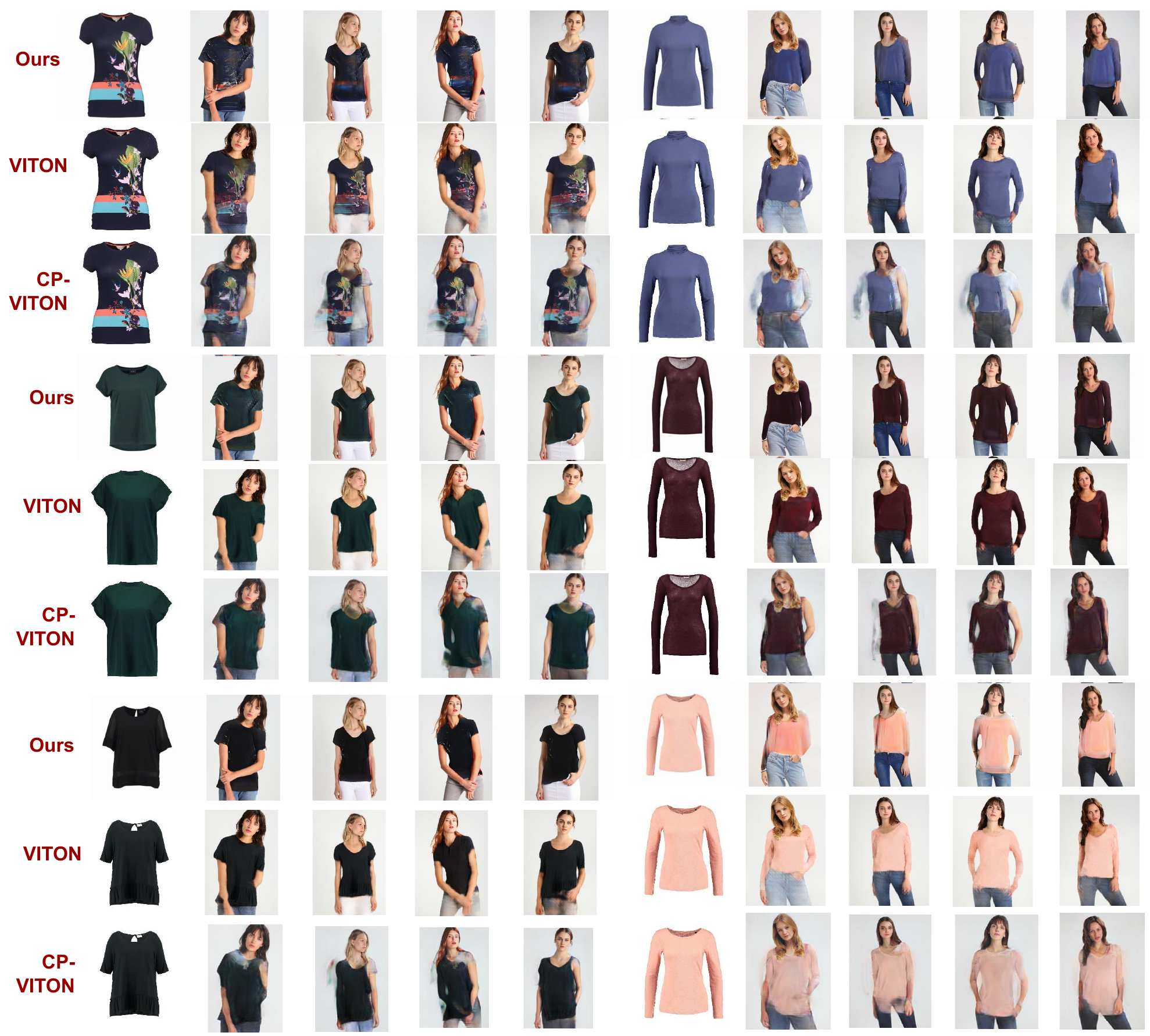}
\caption{
Comparison with VITON (supervised) \cite{han2018viton} on the try-on task.
}
\label{fig:baselineComparsion}
\end{figure}

We compare our model w.r.t. CycleGAN~(\cite{cycleGAN}), MUNIT~(\cite{MUNIT}) and VITON~(\cite{han2018viton}).
Fig.~\ref{fig:baselineComparsion_intro} shows qualitative
results from our model, CycleGAN and MUNIT 
(with/without shared S.E.).
It is clear that these unpaired methods cannot handle the one-to-many shape transfer task. CycleGAN can only work for 
one-to-one mapping task. MUNIT has the ability to do many-to-many mapping for style translation but it is unable to transfer shapes.
We present quantitative results in Table~\ref{tab:quantityBaseline}. We do not provide the Try-on ROI scores for the same reason explained in Sec.~\ref{sec:translationAblationStudy}.
The comparison with the supervised method VITON is shown in Fig.~\ref{fig:baselineComparsion}. It is motivating that even without any supervised paired data, our method achieves competitive results.

\subsection{Ablation Study: Components}
\label{sec:ComponentsAblationStudy}
Given the similarity between MUNIT and our method, 
we ran a more detailed ablation study, see Table~\ref{tab:ablation}, 
in order to analyze the margin between these two methods. 
Note that we use our variant of MUNIT, i.e. MUNIT*, 
where the shape mask is used to segment the features instead of the input image, then each channel dimensions become 1.5 times larger and one more residual block is used in the decoder. 
We also add an additional component ($A$) in the analysis.
When not sampling the style code ($A$ in Table~\ref{tab:ablation}), performance increases significantly. This is an implicit shared style space constraint because there is a style code latent reconstruction loss. When using an explicit shared style encoder ($B$), the performance increases further. In addition, this also enables to extract the compact feature. Further, the Fit-in module ($C$) is essential to enable the try-on stream, since it utilizes the context information guidance to enforce the output to be deterministic.
In our experiments applying multiple AdaIN ($D$) proved useful to stabilize the training process. The mask attention ($E$) and perceptual loss ($F$) further improve the performance. 
Finally, we verified these observations on top of the original MUNIT where introducing $A$ produces a significant increase in performance (as also noted in \cite{MUNIT}), with a further boost when considering $B$.
Note that {\small MUNIT*+A B C} produces blurry images, leading to high SSIM scores~\cite{johnson2016perceptual}


\begin{table}
\setlength{\tabcolsep}{4.7pt} 
\centering
\caption{Comparisons w.r.t. state-of-the-art methods on FashionStyle.
}
\resizebox{\columnwidth}{!}{
\begin{tabular*}{12cm}
{@{\extracolsep{\fill}} l c  }
\toprule 
\multirow{2}{*}{Method} & Take off \\ 
                         & (SSIM/LPIPS-VGG)\\
\midrule[0.6pt]	
CycleGAN  & 45.63 / 47.47   \\
MUNIT  & 45.97 / 46.53    \\
MUNIT, shared S.E. & 50.44 / 49.15 \\
Ours &\textit{\textbf{61.19}} / \textit{\textbf{34.37}} \\
\bottomrule[1pt]
\end{tabular*}
}
%
\label{tab:quantityBaseline}
\end{table}


\begin{table}[h]
\setlength{\tabcolsep}{4.7 pt} 
\centering
\caption{Mean SSIM and LPIPS-VGG similarity when considering the following components: $A$: Use the encoded style code of the input image instead of sampling from the style latent space. $B$: Shared style encoder. $C$: Fit-in module. $D$: Apply AdaIN to the convolutional layers in both Residual and Up-sampling blocks. $E$: Mask attention. $F$: Perceptual loss.
}
\resizebox{\columnwidth}{!}{
\begin{tabular*}{12cm}
{@{\extracolsep{\fill}} l c c }
\toprule 
Method & Try-on ROI  & Take off \\ 
                     & (SSIM/LPIPS) & (SSIM/LPIPS)\\
\midrule[0.6pt]	
MUNIT* & N/A & 55.65 / 43.88 \\ 
MUNIT* + $A$ & N/A & 58.45 / 37.93    \\
MUNIT* + $A$ $B$ & N/A & 58.98 / 36.24   \\
MUNIT* + $A$ $B$ $C$ & \textbf{68.09} / 27.65 & 55.90 / 38.74    \\
MUNIT* + $A$ $B$ $C$ $D$ & 66.48 / 27.70 & 57.21 / 36.83   \\ 
MUNIT* + $A$ $B$ $C$ $D$ $E$ & 66.78 / 27.37  & 58.96 / 36.62   \\
MUNIT* + $A$ $B$ $C$ $D$ $E$ $F$ & 66.42 / \textbf{27.02}& \textbf{61.19} / \textbf{34.37} \\
\bottomrule[0.5pt]
MUNIT & N/A & 51.92 / 48.16   \\
MUNIT + $A$ & N/A & 54.03 / 44.18   \\
MUNIT + $A$ + $B$ & N/A & 55.59 / 43.83 \\
\bottomrule[1pt]
\end{tabular*}
}
\label{tab:ablation}
\end{table}


\subsection{Clothing retrieval}
\label{sec:exp_cloth_retrieval}
We present the in-shop clothing retrieval results using the extracted style features. We apply the shared style encoder as feature extractor to extract the style codes and then use L2 distance to measure the similarity for retrieval. 


\myparagraph{Protocol.} 
The shared style encoder is trained and tested on the FashionStyle training and testing sets, respectively. During retrieval, there are 1,302 query images and 434 database images. 
The query images are all from domain A, i.e. clothed  people. and
database images are from domain B, i.e. individual clothing items.
For fair comparison, we apply the clothing masks to the query input of both our method and other methods.
As shown in Table~\ref{tab:within_retrieval}, we provide four baselines: Color histogram, Autoencoder${+}$GAN (AE+GAN), ResNet-50/152~(\cite{Resnet}) and FashionNet~(\cite{deepFashion}).
Following \cite{Ji2017CrossDomainIR}'s work, we only use the triplet branch of FashionNet. In addition, for the comparison purpose, we use 8 dimensions and 128 dimensions feature by adding one more
fully connected layer after the original one.
For AE+GAN, the latent code of the AE is 128-dimension. We train the
model using both domain A and domain B images. ResNet-50 and 
ResNet-152 are trained from imageNet.

\myparagraph{Discussion.}
Our method outperforms all the baselines except LPIPS-Alex and FashionNet. It is noted that LPIPS-Alex extracts the feature maps of different layers as clothing features, resulting in a very high dimensional feature vector ($\sim$640K dimensions). This costs a lot, both in compute time as well as in storage costs, which both scale linearly with the dimensionality. 
FashionNet is trained in a supervised way and uses a triplet loss.
It is not surprising that its results are better than ours.
Our extracted style code on the other hand has a very low dimension (\eg 8), which can significantly reduce (over 80K times) the computation. 
Furthermore, combining our method with LPIPS-Alex in a simple coarse-to-fine way, \ie first using our method to quickly obtain the coarse top-$k$ results and then using LPIPS-Alex to re-rank these results, can achieve the best performance among the unpaired methods 
while reducing the aforementioned costs significantly.
The $k$ value can be selected as the point where the performance of our method and LPIPS gets close. e.g.
{\scriptsize $k {=} 20$/$k {=} 5$} for Ours ({\scriptsize $SD {=} 8/SD {=} 128$}), or adapted based on user requirements.  
A similar gain in performance can be achieved for the case of the VITON dataset (Table~\ref{tab:VITONretrieval}).

In addition, we provide a clothing retrieval ablation study on FashionStyle, as shown in Fig.~\ref{fig:FashionStyle_Retri_ABL}. It is interesting to observe that the performance of the retrieval process is affected by different factors than that of the image translation process (Sec.~\ref{sec:translationAblationStudy}). 
We hypothesize that the translation task directly exploits shape 
related components in order to achieve detailed image generation. 
On the contrary, the retrieval task considers representative 
features regardless of whether they grant accurate shape transfer.

We also provide the \textit{computation complexity} analysis for the retrieval. 
We use Euclidean distance to measure the difference between the features extracted from two different images.
For each query, computation complexity is {\scriptsize$\mathcal{O}(d \cdot n)$} which scales linearly with the feature dimension $d$ and the number of database images $n$.
Thus, the computation complexity of our method is 80k times ({\scriptsize$SD{=}8$}) or 5k times ({\scriptsize$SD{=}128$}) smaller than LPIPS-Alex according to the dimension in Table~\ref{tab:within_retrieval}. 
As to LPIPS-Alex+Ours, the computation complexity is {\scriptsize $\mathcal{O}(d_{Ours} {\cdot} n + d_{LPIPS} \cdot k), k {\ll} n$} which maintains the performance and significantly reduces the computation compared to {\scriptsize $\mathcal{O}(d_{LPIPS} \cdot n)$,  $d_{Ours} {\ll} d_{LPIPS}$ }
for our naive implementation. While more efficient retrieval algorithms exist, the dependence on the feature dimensionality remains.

\begin{table}
\setlength{\tabcolsep}{4.7pt} 
\centering
\caption{
Retrieval recall rate in the FashionStyle dataset. 
}
\resizebox{\columnwidth}{!}{
\begin{tabular*}{12cm}
{@{\extracolsep{\fill}} l c c c c c}
\toprule 
Method & Dim & top-1 & top-5 & top-20 & top-50 \\
\midrule[0.6pt]	
Color histogram & 128 & 1.5 & 3.7 & 9.3 & 17.4\\
AE+GAN & 128 & 9.4 & 21.9 & 39.6 & 57.3 \\
ResNet-50 (\cite{Resnet}) & 2048 & 11.9 & 25.0 & 40.9 & 56.4 \\
ResNet-152 (\cite{Resnet}) & 2048 & 14.4 & 29.1 & 47.6 & 62.8 \\
LPIPS-Alex (\cite{zhang2018unreasonable}) & 640K & 25.2 & 42.0 & 59.5 & 72.0 \\
FashionNet (\textit{D=4096}) (\cite{deepFashion}) & 4096 & 30.0 & 56.2  & 79.3  & 90.9 \\ 
FashionNet (\textit{D=128}) (\cite{deepFashion}) & 128 & 26.3 & 52.3  & 78.0  & 89.6 \\ 
FashionNet (\textit{D=8}) (\cite{deepFashion}) & 8 & 19.4 & 47.0  & 74.5  & 88.2 \\ 
\midrule[0.6pt]	
\textit{Ours (SD = 8)} & 8 & 17.1 & 37.6 & 58.1 & 74.3 \\
\textit{Ours (SD = 32)} & 32 & 18.7 & 39.6 & 62.5 & 76.1 \\
\textit{Ours (SD = 128)} & 128 & 19.4 & 41.1 & 64.1 & 77.6 \\
\textit{LPIPS-Alex + Ours(SD = 8, k = 20)} & - & 24.4 & 41.4 & 58.1 & 74.3 \\
\textit{LPIPS-Alex + Ours(SD = 128, k = 5)} & - & 24.4 & 41.1 & 64.1 & 77.6 \\
\bottomrule[1pt]
\end{tabular*}
}
\label{tab:within_retrieval}
\end{table}

\begin{figure}
\centering
\includegraphics[width=0.5\textwidth]{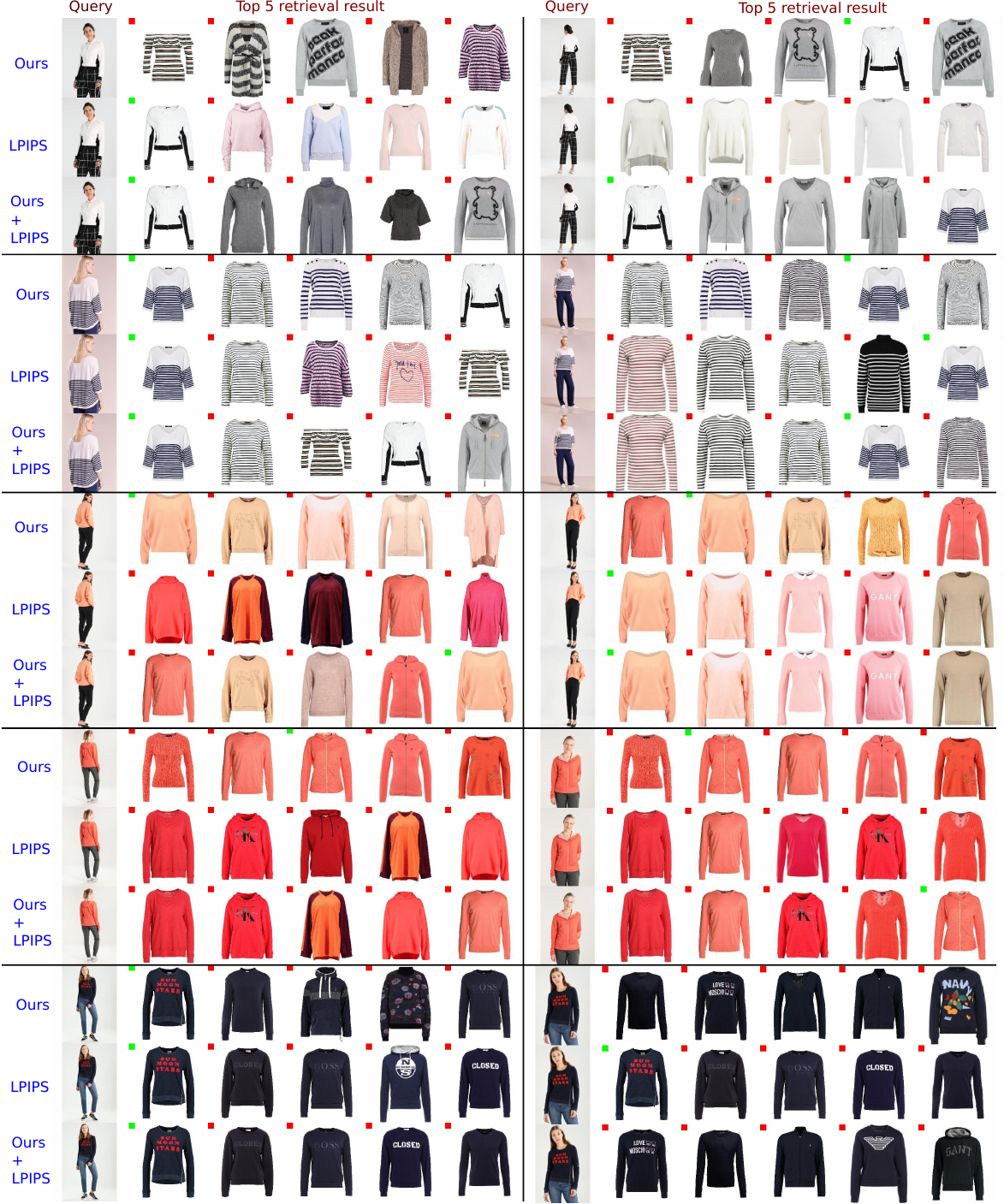}
\caption{
Top-5 retrieval results on the FashionStyle dataset sorted in 
decreasing order from left to right.  
Correct items are marked in green.
}
\label{fig:FashionStyleRetrieval}
\end{figure}


\begin{table}
\setlength{\tabcolsep}{3pt} 
\centering
\caption{Retrieval recall rate in the VITON dataset}
\resizebox{1\columnwidth}{!}{
\begin{tabular*}{12cm}
{@{\extracolsep{\fill}} l c c c c c}
\toprule 
Method & Dim & top-1 & top-5 & top-20 & top-50 \\
\midrule[0.6pt]	
Ours & 128 & 20.2 & 39.9 & 64.9 & 79.3\\
LPIPS-Alex (\cite{zhang2018unreasonable})& 640K & 42.3 & 61.3 & 77.2 & 88.7 \\
LPIPS-Alex + Ours(k = 50) & -& 41.6 & 57.7 & 72.8 & 79.3 \\
\bottomrule[1pt]
\end{tabular*}
}
\label{tab:VITONretrieval}
\end{table}


\begin{table}
\setlength{\tabcolsep}{4.7pt} 
\centering
\caption{Mean SSIM and LPIPS-VGG distance of face experiment.}
\resizebox{1\columnwidth}{!}{
\begin{tabular*}{12cm}
{@{\extracolsep{\fill}} l c c }
\toprule 
\multirow{2}{*}{Method} & Try-on ROI  & Take off \\ 
                        & (SSIM/LPIPS-VGG) & (SSIM/LPIPS-VGG)\\

\midrule[0.6pt]	
Face experiment & 69.48 / 15.65  & 43.82 / 39.97   \\

\bottomrule[1pt]
\end{tabular*}
}
\label{tab:FACES_similarity}
\end{table}

\subsection{Face shape transfer}
\label{sec:exp_face}

\begin{figure}[h]
\centering
\includegraphics[width=0.5\textwidth]{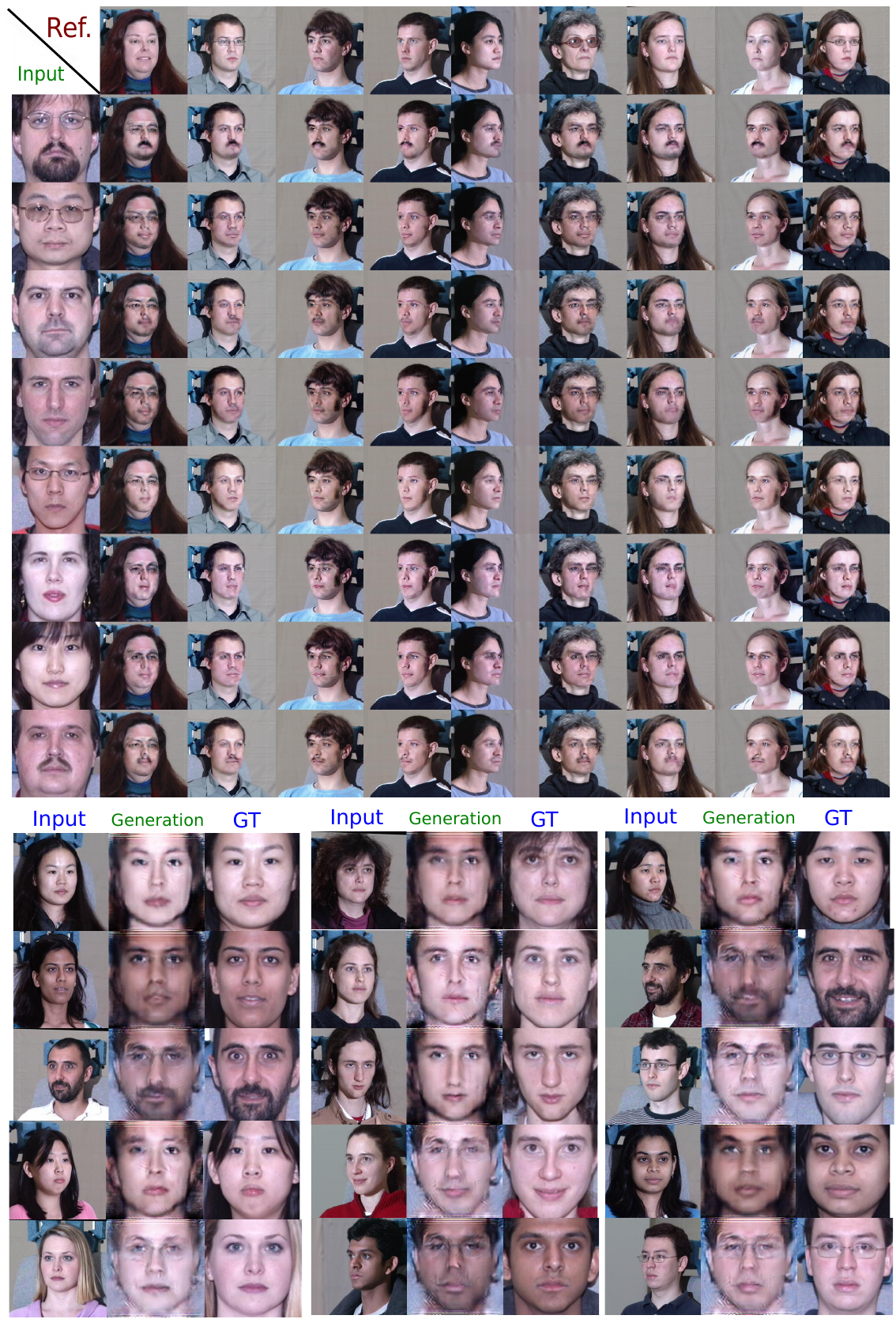}

\caption{
Face translation. Top, given the input face and the target body, we generate a new image where the input face is fitted to the target body (try-on), and vice versa (take-off) at the bottom.
}
\label{fig:faceQuality}
\end{figure}

We conduct experiments related to face translation. 
In the first experiment, given the input face and the target
context (body), we generate a new image where the input face 
is fitted on the target context (try-on task). 
In the second experiment, we perform a face take-off task 
where given a face image with a side viewpoint, we generate an image where the face from the input is rotated towards the front and zoomed-in.
We conduct these experiments on the CMU MultiPIE dataset.
Qualitative results are presented in Fig.~\ref{fig:faceQuality}.
%
We present translation similarity measurements 
in Table~\ref{tab:FACES_similarity}.

\myparagraph{Discussion}
As can be noted in Fig.~\ref{fig:faceQuality}, images from the different domains, i.e. frontal 
and side view faces, exhibit many differences regarding to scale and the presence of other parts of the body. Yet, the proposed method is able to achieve both translation tasks with a decent level of success.
Fig.~\ref{fig:faceQuality} shows that, for both tasks, apart from facial orientation features such as facial hair, lip color, accessories, and skin color are to some level properly translated. It is remarkable that this has been achieved without using 
facial landmarks like eyes, nose, mouth, ears, as in existing work~(\cite{huang2017beyond,Zhao_2018_CVPR}).
Failures are mainly caused by incorrectly estimated masks, large pose variation, and inconsistent skin colors\footnotemark[2].
Table~\ref{tab:FACES_similarity} shows that the proposed method has a comparable performance on both faces and clothing related datasets.

\section{Conclusion}
\label{sec:conclusion}
We present a method to translate the shape of an object
across different domains. 
%
Extensive empirical evidence suggests that our method has comparable performance on both faces and clothing data.
Moreover, our ablation study shows that the proposed 
mask attention and Fit-in module assist the translation 
of shape, thus, improving the generation process.
Finally, we have shown that the features learned 
by the model have the potential to be employed for 
retrieval tasks, in spite of their low dimensionality.

\vspace{4mm}

\noindent\textit{Acknowledgements:} 
This work was funded by the Agentschap Innoveren en Ondernemen (VLAIO) project HBC.2017.0358.

{\small
\bibliographystyle{ieee_fullname}
\bibliography{egbib}

\begin{thebibliography}{10}\itemsep=-1pt

\bibitem{AugmentedCycleGAN}
Amjad Almahairi, Sai Rajeswar, Alessandro Sordoni, Philip Bachman, and Aaron
  Courville.
\newblock Augmented cyclegan: Learning many-to-many mappings from unpaired
  data.
\newblock In {\em ICML}, 2018.

\bibitem{anonymous2019unsupervised}
Anonymous.
\newblock Unsupervised one-to-many image translation.
\newblock In {\em Submitted to International Conference on Learning
  Representations}, 2019.
\newblock under review.

\bibitem{WGAN}
Mart{\'{\i}}n Arjovsky, Soumith Chintala, and L{\'{e}}on Bottou.
\newblock Wasserstein generative adversarial networks.
\newblock In {\em ICML}, 2017.

\bibitem{UnseenPose}
Guha Balakrishnan, Amy Zhao, Adrian~V Dalca, Fredo Durand, and John Guttag.
\newblock Synthesizing images of humans in unseen poses.
\newblock In {\em CVPR}, 2018.

\bibitem{ColorGAN}
Yun Cao, Zhiming Zhou, Weinan Zhang, and Yong Yu.
\newblock Unsupervised diverse colorization via generative adversarial
  networks.
\newblock In {\em {ECML/PKDD}}, 2017.

\bibitem{cao2017realtime}
Zhe Cao, Tomas Simon, Shih-En Wei, and Yaser Sheikh.
\newblock Realtime multi-person 2d pose estimation using part affinity fields.
\newblock In {\em CVPR}, 2017.

\bibitem{GAN4augmentation}
M. Frid-Adar, E. Klang, M. Amitai, J. Goldberger, and H. Greenspan.
\newblock Synthetic data augmentation using gan for improved liver lesion
  classification.
\newblock In {\em ISBI}, 2018.

\bibitem{gatys2016image}
Leon~A Gatys, Alexander~S Ecker, and Matthias Bethge.
\newblock Image style transfer using convolutional neural networks.
\newblock In {\em CVPR}, 2016.

\bibitem{GAN}
Ian~J. Goodfellow, Jean Pouget{-}Abadie, Mehdi Mirza, Bing Xu, David
  Warde{-}Farley, Sherjil Ozair, Aaron~C. Courville, and Yoshua Bengio.
\newblock Generative adversarial nets.
\newblock In {\em NIPS}, 2014.

\bibitem{MultiPIE}
R. Gross, I. Matthews, J. Cohn, T. Kanade, and S. Baker.
\newblock Multi-pie.
\newblock In {\em 2008 8th IEEE International Conference on Automatic Face
  Gesture Recognition}, 2008.

\bibitem{sketchBasedRetrieval}
Longteng Guo, Jing Liu, Yuhang Wang, Zhonghua Luo, Wei Wen, and Hanqing Lu.
\newblock Sketch-based image retrieval using generative adversarial networks.
\newblock In {\em ACM Multimedia}, 2017.

\bibitem{han2018viton}
Xintong Han, Zuxuan Wu, Zhe Wu, Ruichi Yu, and Larry~S Davis.
\newblock Viton: An image-based virtual try-on network.
\newblock In {\em CVPR}, 2018.

\bibitem{Resnet}
Kaiming He, Xiangyu Zhang, Shaoqing Ren, and Jian Sun.
\newblock Deep residual learning for image recognition.
\newblock {\em arXiv preprint arXiv:1512.03385}, 2015.

\bibitem{huang2017beyond}
Rui Huang, Shu Zhang, Tianyu Li, Ran He, et~al.
\newblock Beyond face rotation: Global and local perception gan for
  photorealistic and identity preserving frontal view synthesis.
\newblock In {\em ICCV}, 2017.

\bibitem{huang2017arbitrary}
Xun Huang and Serge~J Belongie.
\newblock Arbitrary style transfer in real-time with adaptive instance
  normalization.
\newblock In {\em ICCV}, 2017.

\bibitem{MUNIT}
Xun Huang, Ming Liu, Serge Belongie, and Jan Kautz.
\newblock Multimodal unsupervised image-to-image translation.
\newblock In {\em ECCV}, 2018.

\bibitem{pix2pix}
Phillip Isola, Jun{-}Yan Zhu, Tinghui Zhou, and Alexei~A. Efros.
\newblock Image-to-image translation with conditional adversarial networks.
\newblock In {\em CVPR}, 2017.

\bibitem{spatialTransformerNet}
Max Jaderberg, Karen Simonyan, Andrew Zisserman, and koray kavukcuoglu.
\newblock Spatial transformer networks.
\newblock In {\em NIPS}. 2015.

\bibitem{Ji2017CrossDomainIR}
Xin Ji, Wei Wang, Meihui Zhang, and Yang Yang.
\newblock Cross-domain image retrieval with attention modeling.
\newblock In {\em ACM Multimedia}, 2017.

\bibitem{neuralStyleTransfer}
Yongcheng Jing, Yezhou Yang, Zunlei Feng, Jingwen Ye, and Mingli Song.
\newblock Neural style transfer: {A} review.
\newblock {\em CoRR}, abs/1705.04058, 2017.

\bibitem{johnson2016perceptual}
Justin Johnson, Alexandre Alahi, and Li Fei-Fei.
\newblock Perceptual losses for real-time style transfer and super-resolution.
\newblock In {\em ECCV}, 2016.

\bibitem{Adam}
Diederik Kingma and Jimmy Ba.
\newblock Adam: A method for stochastic optimization.
\newblock In {\em ICLR}, 2015.

\bibitem{SRGAN}
Christian Ledig, Lucas Theis, Ferenc Huszar, Jose Caballero, Andrew~P. Aitken,
  Alykhan Tejani, Johannes Totz, Zehan Wang, and Wenzhe Shi.
\newblock Photo-realistic single image super-resolution using a generative
  adversarial.
\newblock In {\em CVPR}, 2015.

\bibitem{leeContextAwareSynthesis}
Donghoon Lee, Sifei Liu, Jinwei Gu, Ming-Yu Liu, Ming-Hsuan Yang, and Jan
  Kautz.
\newblock Context-aware synthesis and placement of object instances.
\newblock In {\em NIPS}, 2018.

\bibitem{lee2018diverse}
Hsin-Ying Lee, Hung-Yu Tseng, Jia-Bin Huang, Maneesh Singh, and Ming-Hsuan
  Yang.
\newblock Diverse image-to-image translation via disentangled representations.
\newblock In {\em ECCV}, 2018.

\bibitem{liang2018look}
Xiaodan Liang, Ke Gong, Xiaohui Shen, and Liang Lin.
\newblock Look into person: Joint body parsing \& pose estimation network and a
  new benchmark.
\newblock {\em TPAMI}, 2018.

\bibitem{lin2018stgan}
Chen-Hsuan Lin, Ersin Yumer, Oliver Wang, Eli Shechtman, and Simon Lucey.
\newblock St-gan: Spatial transformer generative adversarial networks for image
  compositing.
\newblock In {\em CVPR}, 2018.

\bibitem{trafficScenesGAN}
G. Liu, J. Wang, C. Zhang, S. Liao, and Y. Liu.
\newblock Realistic view synthesis of a structured traffic environment via
  adversarial training.
\newblock In {\em CAC}, 2017.

\bibitem{UNIT}
Ming-Yu Liu, Thomas Breuel, and Jan Kautz.
\newblock Unsupervised image-to-image translation networks.
\newblock In {\em NIPS}, 2017.

\bibitem{deepFashion}
Ziwei Liu, Ping Luo, Shi Qiu, Xiaogang Wang, and Xiaoou Tang.
\newblock Deepfashion: Powering robust clothes recognition and retrieval with
  rich annotations.
\newblock In {\em CVPR}, 2016.

\bibitem{ma2018exemplar}
Liqian Ma, Xu Jia, Stamatios Georgoulis, Tinne Tuytelaars, and Luc Van~Gool.
\newblock Exemplar guided unsupervised image-to-image translation with semantic
  consistency.
\newblock {\em ICLR}, 2019.

\bibitem{ma2017pose}
Liqian Ma, Xu Jia, Qianru Sun, Bernt Schiele, Tinne Tuytelaars, and Luc
  Van~Gool.
\newblock Pose guided person image generation.
\newblock In {\em NIPS}, 2017.

\bibitem{ma2017disentangled}
Liqian Ma, Qianru Sun, Stamatios Georgoulis, Luc Van~Gool, Bernt Schiele, and
  Mario Fritz.
\newblock Disentangled person image generation.
\newblock In {\em CVPR}, 2018.

\bibitem{LSGAN}
Xudong Mao, Qing Li, Haoran Xie, Raymond~YK Lau, Zhen Wang, and Stephen~Paul
  Smolley.
\newblock Least squares generative adversarial networks.
\newblock In {\em ICCV}, 2017.

\bibitem{DCGAN}
Alec Radford, Luke Metz, and Soumith Chintala.
\newblock Unsupervised representation learning with deep convolutional
  generative adversarial networks.
\newblock In {\em ICLR}, 2016.

\bibitem{SwapNet}
Amit Raj, Patsorn Sangkloy, Huiwen Chang, Jingwan Lu, Duygu Ceylan, and James
  Hays.
\newblock Swapnet: Garment transfer in single view images.
\newblock In {\em ECCV}, 2018.

\bibitem{VGG}
Karen Simonyan and Andrew Zisserman.
\newblock Very deep convolutional networks for large-scale image recognition.
\newblock In {\em ICLR}, 2015.

\bibitem{wang2018toward}
Bochao Wang, Huabin Zheng, Xiaodan Liang, Yimin Chen, Liang Lin, and Meng Yang.
\newblock Toward characteristic-preserving image-based virtual try-on network.
\newblock In {\em ECCV}, 2018.

\bibitem{wangICIP20}
Kaili Wang, Liqian Ma, Jose Oramas~M., Luc Van~Gool, and Tinne Tuytelaars.
\newblock Unpaired image-to-image shape translation across fashion data.
\newblock In {\em ICIP}, 2020.

\bibitem{SSIM}
Zhou Wang, A.~C. Bovik, H.~R. Sheikh, and E.~P. Simoncelli.
\newblock Image quality assessment: From error visibility to structural
  similarity.
\newblock {\em Trans. Img. Proc.}, 2004.

\bibitem{wu2018light}
Xiang Wu, Ran He, Zhenan Sun, and Tieniu Tan.
\newblock A light cnn for deep face representation with noisy labels.
\newblock {\em IEEE Trans. on Information Forensics and Security}, 2018.

\bibitem{YooPixelLevelTransfer16}
Donggeun Yoo, Namil Kim, Sunggyun Park, Anthony~S. Paek, and In{-}So Kweon.
\newblock Pixel-level domain transfer.
\newblock In {\em ECCV}, 2016.

\bibitem{zhang2018unreasonable}
Richard Zhang, Phillip Isola, Alexei~A Efros, Eli Shechtman, and Oliver Wang.
\newblock The unreasonable effectiveness of deep features as a perceptual
  metric.
\newblock In {\em CVPR}, 2018.

\bibitem{Zhao_2018_CVPR}
Jian Zhao, Yu Cheng, Yan Xu, Lin Xiong, Jianshu Li, Fang Zhao, Karlekar
  Jayashree, Sugiri Pranata, Shengmei Shen, Junliang Xing, Shuicheng Yan, and
  Jiashi Feng.
\newblock Towards pose invariant face recognition in the wild.
\newblock In {\em CVPR}, 2018.

\bibitem{T2Net}
Chuanxia Zheng, Tat{-}Jen Cham, and Jianfei Cai.
\newblock T {\^{}}2 2 net: Synthetic-to-realistic translation for solving
  single-image depth estimation tasks.
\newblock In {\em ECCV}, 2018.

\bibitem{cycleGAN}
Jun{-}Yan Zhu, Taesung Park, Phillip Isola, and Alexei~A. Efros.
\newblock Unpaired image-to-image translation using cycle-consistent
  adversarial networks.
\newblock In {\em ICCV}, 2017.

\end{thebibliography}
}

\end{document}